\documentclass[final]{cvpr}

\usepackage{times}
\usepackage{epsfig}
\usepackage{graphicx}
\usepackage{amsmath}
\usepackage{amssymb}

\usepackage{url}
\usepackage{subfigure}
\usepackage{amsmath}
\usepackage{xcolor}
\usepackage{adjustbox}
\usepackage[bottom]{footmisc}
\usepackage{mathtools,amsmath}
\usepackage{amsthm}
\DeclareMathOperator*{\argmax}{arg\,max}
\DeclareMathOperator*{\argmin}{arg\,min}
\newcommand{\norm}[1]{\lVert#1\rVert}

\newcommand{\WGAP}[2]{WGAP-J\textsubscript{#1}($\epsilon_{#2}$) }
\newcommand{\GAP}[1]{GAP($\epsilon_{#1}$)}

\usepackage{floatrow}
\DeclareFloatFont{tiny}{\tiny}

\usepackage{booktabs}

\usepackage[pagebackref=true,breaklinks=true,colorlinks,bookmarks=false]{hyperref}

\def\cvprPaperID{8751} 
\def\confYear{CVPR 2021}

\usepackage{graphicx}
\graphicspath{
{{C:/projects/CVPR2021/ImageResults/}}
}

\begin{document}

\title{Adversarial Attacks with Time-Scale Representations}

\author{Alberto Santamaria-Pang, Jianwei Qiu, Aritra Chowdhury, James Kubricht,\\
Peter Tu, Iyer Naresh, Nurali Virani\\
GE Research, 1 Research Circle, Niskayuna, NY 12309\\
{\tt\small \{santamar, jianwei.qiu, Aritra.Chowdhury,
james.kubrich, tu, iyerna, nurali.virani\}@ge.com}
}
\maketitle

\begin{abstract}
We propose a novel framework for real-time black-box universal attacks which disrupts activations
of early convolutional layers in deep learning models. Our hypothesis is that perturbations produced
in the wavelet space disrupt early convolutional layers more effectively than perturbations performed
in the time domain. The main challenge in adversarial attacks is to preserve low frequency image content
while minimally changing the most meaningful high frequency content. To address this, we formulate
an optimization problem using time-scale (wavelet) representations as a dual space in three steps.
First, we project original images into orthonormal sub-spaces for low and high scales via wavelet
coefficients. Second, we perturb wavelet coefficients for high scale projection using a generator network.
Third, we generate new adversarial images by projecting back the original coefficients from the
low scale and the perturbed coefficients from the high scale sub-space.
We provide a theoretical framework that guarantees a dual mapping from time and time-scale domain representations.
We compare our results with state-of-the-art black-box attacks from generative-based and gradient-based models. 
We also verify efficacy
against multiple defense methods such as JPEG compression, Guided Denoiser and Comdefend.
Our results show that wavelet-based perturbations consistently outperform time-based attacks
thus providing new insights into vulnerabilities of deep learning models and could potentially lead to robust
architectures or new defense and attack mechanisms by leveraging time-scale representations.
\end{abstract}

\section{Introduction}

\begin{figure}[t]
  \begin{center}
  \includegraphics[width=0.99\linewidth]{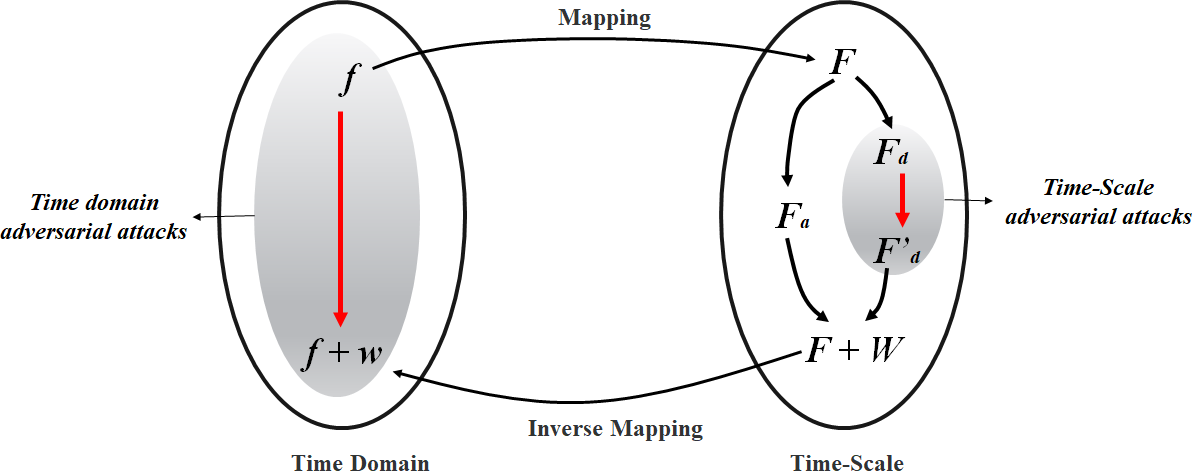}
  \end{center}
   \caption{Schematic of the proposed Time-Scale adversarial attack.
   }
    \label{Method01}
\end{figure}

While great progress has been made in adversarial attacks~\cite{brunner_guessing_2019,chen_zoo_2017,ilyas_prior_2019,tsipras_robustness_2019},
the fundamental reason why convolutional neural networks (CNNs) are fragile to small perturbations
 is still unclear~\cite{moosavi-dezfooli_universal_2017,samangouei_defense-gan_2018,carlini_evaluating_2019,wong_provable_nodate,salem_ml-leaks_2018}.
State-of-the-art models generate perturbations that seem like statistical noise that affect both high and low image scales (similar to high and low frequency).
We believe that the inability to localize an attack becomes a key limitation of
time-domain attacks as they end up modifying unnecessary or unrelated information in the image~\cite{li_adversarial_2017,borji_pros_2019}.
Thus,  instead of generating adversarial attacks in the time domain (low and high frequencies),
we propose to expose a new vulnerability of deep learning models by generating fine structural-perturbations in meaningful image content in the high scale Wavelet decomposition.
Studies have shown that early convolution layers capture low-level edge and edge orientation
information possibly similar to receptive fields of human vision~\cite{kheradpisheh_deep_2016,zeiler_visualizing_2014}.
We aim to induce a perturbation that will alter the local vertical, horizontal, and diagonal edge content of images, therefore,
creating arbitrary activation spikes in early convolution layers. We argue that decomposing and isolating localized coherent
spatial arrangements is an effective mechanism for disrupting CNN layer responses. Our approach relies on the following steps:
i) time-scale analysis is used to provide a mechanism to decompose an image and localize high scale (frequency) and structural image properties,
ii) generative models are used to learn both high-scale and structural patterns to perturb high scale coefficients,
iii) synthesis of image from perturbed coefficients aims disrupt early convolution layers. Figure~\ref{Method01} presents a schematic of our
proposed method.

Since,  state-of-the-art deep learning (DL) network designs rely on convolutional layers that broadly target
all high frequency content -- indiscriminate to structural perturbations -- making them highly
vulnerable to these attacks~\cite{sarkar_upset_2017,ferrari_practical_2018,vargas_robustness_2019,xie_adversarial_2017,carlini_adversarial_2017}.
The targeted decomposition of fine image details accounts for a small fraction of the image energy content~\cite{mallat_understanding_2016,heil_homogeneous_2007,daubechies_wavelet_1990}.
This helps with our objective to minimally perturb the image while inducing misclassification in classifier that consist of convolution layers.
The significance of our approach is that the concept of scales (wavelet decomposition levels) and orientations (filter banks)
 can be associated to one another to investigate the fundamental principles of image/video analysis and signal processing for a new type of black-box attack.
\section{Literature Review}
\label{SecLitReview}
To the best of our knowledge, no work has been reported in attack-based mechanisms from time-scale representations.
All adversarial attack frameworks reported to date~\cite{prakash_deflecting_2018,li_regional_2019,sharif_general_2019,mopuri_fast_2017,guo_low_2019,cheng_improving_2019,chen_zoo_2017,su_one_2019,xiao_advit_2019,joshi_semantic_2019,ferrari_gray-box_2018},
perturb an image which is represented in the time domain using all of its frequency (Fourier) content. A detail review can be found in Pitropakis~\textit{et al.}~\cite{pitropakis_taxonomy_2019}.
Black-box attacks are challenging adversarial attacks as there is
no access to training data nor to the target model.
These attacks can be categorized as either \textit{gradient-based} or \textit{transfer-based} methods. \textit{Gradient-based}
methods iteratively estimate the gradient by perturbing the input image while optimizing an objective function.
Here the major challenge is to account for robust optimization in a high dimensional
input space (see Chen~\textit{et al.}~\cite{chen_zoo_2017} for an initial report). Ilyas~\textit{et al.}~\cite{ilyas_prior_2019}
subsequently proposed a natural evolution attack with a limited number of queries.
While their methods showed a significant improvement in query efficiency and speed,
it generated images with high distortion. Performance was later improved by Cheng~\textit{et al.}~\cite{cheng_improving_2019}
who used a surrogate model as a prior, effectively reducing the number of gradient averages or queries per iteration.

To alleviate computational cost, \textit{transfer-based} attacks~\cite{poursaeed_generative_2018,shafahi_adversarial_2019,shafahi_universal_2019,hayes_learning_2018}
train a generator network using white-box attacks from a surrogate or a chosen base model. Sharif~\textit{et al.}~\cite{shafahi_adversarial_2019} used
adversarial generative networks to create adversarial examples with defined objectives.
The method was demonstrated in applications for face and handwritten-digit recognition. Poursaeed~\textit{et al.}~\cite{poursaeed_generative_2018}
presented an approach that uses a generative model to design small perturbations (high frequency)
that are distributed in the overall image, rather than following structural patterns.
All of the above methods share the following limitations: \textit{gradient-based} methods are prohibitively
computationally expensive when estimating the gradient, while \textit{transfer-based} methods suffer from generalization.
\textit{Our hypothesis} is that accounting for unnecessary and redundant image content
in very high multidimensional space leads to non-tractable optimization problems and poor generalization~\cite{grosse_statistical_2017,vargas_robustness_2019,ferrari_characterizing_2018,mopuri_generalizable_2019,stutz_disentangling_2019}.

\textit{Our main contributions} are the following: i) providing a novel, tractable, fast, and black-box (surrogate-based or transfer-based) adversarial attack for deep neural networks, ii) exploring a new direction to study vulnerability (and hence robustness) of neural networks via time-scale perturbations that affects early convolution layers, and iii) experimental study with state-of-the-art classifiers and other generative or iterative attacks to illustrate the efficacy and salient features of this attack. Our code is available online at:
\text{https://github.com/waveletgap/wgap}.


\section{Methods}
\label{SecMethods}

\subsection{Background}

\begin{figure}[t]
	\begin{center}
		\includegraphics[width=0.9\linewidth]{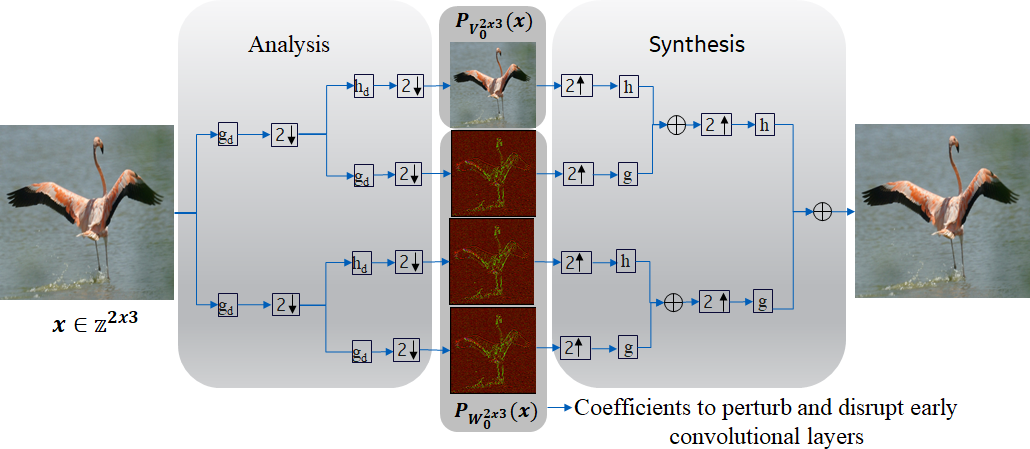}
	\end{center}
    \vspace{-0.4cm}
	\caption{Schematic of the 2D wavelet transform for an RGB image $x$ showing associated filter banks for analysis and synthesis. 
    The input image is projected into $P_{V_0^{2\times3}}$  and  $P_{W_0^{2\times3}}$. }
	\label{FigWavelet}
\end{figure}

\paragraph{Multiresolution Analysis:} We briefly review multiresolution analysis introduced by Mallat and Meyer from~\cite{mallat_wavelet_nodate}. Let $\varphi$ a function in $L^2(\mathbb{R})$ (Hilbert space of square-integrable complex valued functions on $\mathbb{R}$), such that the translation $\varphi(\cdot - k), k \in \mathbb{Z}$ defines an orthonormal system. Then $\varphi$ generates the sequence of spaces $V_j : j \in \mathbb{N}$, if $\varphi_{jk}(x) = 2^{j/2} \varphi(2^j x - k   )$.
A multiresolution analysis (MRA) for $L^2(\mathbb{R})$ is a sequence of spaces $\{V_j\}$ generated by the orthonormal system $\varphi_{jk}$ in $L^2(\mathbb{R})$, if $ V_j \subset V_{j+1}$ and $ \bigcup V_j $ is dense in $L^2(\mathbb{R})$.  We say that $\varphi$ is a \textit{father wavelet} or scaling function if $\varphi$ generates a MRA. Given $ f \in L^2(\mathbb{R}) $, it can be shown that $f \in \bigcup_{j=0}^{\infty} V_{j} = V_{0} \oplus \bigoplus_{j=0}^{\infty} W_j $, where $W_j$ is the orthogonal complement of $V_{j}$ in $V_{j+1}$. We refer to $W_j$ as $j^{th}$ \textit{resolution level} of the multiresolution analysis of level $j$. A \textit{mother wavelet} is a function $\psi \in W_0 $ such that $\psi_{0k} = \psi(\cdot - k)$. Then $\psi$ is an orthonormal basis in $W_j$ and is orthogonal to the father wavelet. Then any $f \in L^2(\mathbb{R})$  can be represented in terms of the father and mother wavelet via the following series:
\begin{equation}\label{waveletSeries}
f(x)=\sum_k\alpha_k \varphi_{0k}(x) + \sum_{j=0}^\infty \sum_k \beta_{jk} \psi_{jk} (x),
\end{equation}
where the wavelet coefficients are defined as: $\alpha_k = \int f(x) \overline{ \varphi_{0k}(x)} dx,~~\beta_{jk} =  \int f(x) \overline{\psi_{jk} (x)} dx$.
The main advantage of the wavelet representation in Eq.~\ref{waveletSeries} over  Fourier analysis is the ability to capture both the \textit{spatial component} and the \textit{frequency spectrum}. The term $\sum_k\alpha_k \varphi_{0k}(x)$ is the \textit{spatial component} whereas the term $\sum_{j=0}^\infty \sum_k \beta_{jk} \psi_{jk} (x)$ localizes the \textit{frequency spectrum}. The coefficients $\alpha_k$ inherently capture the main signal content while $\beta_{jk}$ capture the local details. We refer to $\alpha_k$ and  $\beta_{jk}$ as the approximation and detail coefficients, respectively.
\paragraph{2D Discrete Wavelet Transform (DWT):} The 2D Discrete Wavelet Transform (DWT) is a direct generalization of the 1D MRA into $L^2(\mathbb{Z}^2)$. Let $\varphi$ be a scaling function and  $\psi$ its corresponding wavelet. We define three wavelets: $\psi^1 = \varphi \psi$, $\psi^2 = \psi \varphi$, and $\psi^3 = \varphi \varphi$, where $\psi_{j,(n_1,n_2)}^k(t_1,t_2) = 2^{j/2} \psi^k(( 2^j n_1 - t_1)/2^j, ( 2^j n_2 - t_2)/2^j)$ with $k\in\{1,2,3\}$. Then the wavelet family $\{ \psi_{j,n}^1, \psi_{j,n}^2, \psi_{j,n}^3 \}_{n \in Z^2}$ with $n = ( n_1, n_2)  $ is an orthonormal basis for $W_j^2$  and $\{ \psi_{j,n}^1, \psi_{j,n}^2, \psi_{j,n}^3  \}_{j,n \in Z^2}$ is an orthonormal basis for $L^2(\mathbb{Z}^2)$. The corresponding coefficients are obtained from the inner products: $a_j[n] = \langle f, \varphi_{j,n}^2\rangle ,~ d_{jk}[n] = \langle f, \psi_{j,n}^k\rangle,\; \forall k \in \{1,2,3\}$. The approximation coefficients $a_j$ are associated with low frequencies, whereas the detail coefficients $d_{jk}$ are associated with high frequencies at horizontal, vertical, and diagonal orientations. The scaling function $\varphi$ and the mother wavelet $\psi$ can be represented as conjugate mirror filters banks $h[n]$ and $g[n]$. We denote as $h_d[n]=h[-n]$ and $g_d[n]=[-g]$ the mirror filters of $h[n]$ and $g[n]$, respectively.
Figure~\ref{FigWavelet} illustrates two-dimensional DWT for the first decomposition level.
\subsection{Wavelet-based Generative Adversarial Perturbation (WGAP)  Modeling}
We present a generic framework for black-box and transfer attacks by perturbing the projection $P_{W_{j_0}}(x)$ (details)
and preserving the projection $P_{V_{j_0}}(x)$ (approximation) for an image $x$. The analysis is shown for single channel images $L^2(\mathbb{Z}^2)$, but is easily generalized to three-channel color images. Since it has been observed that early convolution layers capture low-level edge and edge orientation information, we aim to induce a perturbation in $P_{W_{j_0}}(x)$ such that it will alter the local vertical, horizontal, and diagonal edge content of $x$, therefore, creating arbitrary activation spikes in early convolution layers. This method systematically alters the edges, thereby influencing shape-content information. This approach is different from denoising since noise distribution is not explicitly modeled. Rather, object structure from finer details is directly altered. We also deviate from intervention methods in the high-frequency Fourier spectrum since Fourier transformations lose spatial content coherence in contrast with time- and space-scale representations used in this work.

\subsubsection{General Formulation}
Suppose, we have a classiﬁer $c(\cdot)$ with a loss function $\ell_c(X,y)$ where $X$ and $y$ is the input image and output class label, respectively. Here $c(X)$ is the predicted label. We assume: i) an adversarial example can be derived from an additive model as follows $X' = X + D$, and ii) for a fixed class label $y$, the loss $\ell_c$ is a continuous function. Then adversarial attack generation can be posed as an optimization problem:
\begin{equation}\label{adversarialMax}
\argmax_{X' : \norm{X'-X} \le \epsilon} \ell_c\big(X',c(X)\big).
\end{equation}
Please note that: i) if $c(\cdot)$ is the attacked model then it is a white-box attack,
ii) if it is a surrogate of the model to be attacked, then it is a black-box attack,
and iii) if it is a chosen base model and the model to be attacked is unknown, then it is a transfer attack.
Based on this formulation, in this work, we are interested in transfer and black-box attacks
with perturbation in the wavelet domain.
To convince that such an attack is sound, we look at the following result.

\textbf{Theorem 1:}  Let $( a_j,d_{jk} )$, $( a_j,d'_{jk} )$ be wavelet coefficients for $X$ and $X'$,  such that
$X=\text{DWT}^{-1}(a_{j},d_{jk})$ and $X'=\text{DWT}^{-1}(a_{j},d'_{jk})$.
For some scale $j_0$, we can express problem in Equation~\ref{adversarialMax} in terms of the wavelet coefficients as follows:
\begin{equation}\label{adversarialMaxWavelet}
\argmax_{\text{DWT}^{-1}(a_{j_0}, d'_{j_0k}) :  \norm{ d_{j_0k} - d'_{j_0k}} \le \tilde{\epsilon}} \ell_c\big(X',c(X)\big).
\end{equation}
\begin{proof}
We need to prove that if $\norm{X'-X} < \epsilon$, then there exists $\tilde{\epsilon}$, such that $\norm{ d_{j_0k}-d'_{j_0k}} < \tilde{\epsilon}$ for any chosen scale $j_0$.
Given that $X$ and $X'$ have the same approximation coefficients, from Equation~\ref{waveletSeries},
$D= X' - X $ is in $W_j^2$ but not in $V_0$. Then using, linearity, orthonormality and the  Pythagorean identity, we can write $\norm{X-X'}$ as:
\begin{equation}
\norm{\sum_{j=0}^\infty \sum_k d_{jk} \psi_{jk} (X) - \sum_{j=0}^\infty \sum_k d'_{jk} \psi_{jk} (X')} =
\end{equation}
\begin{equation}
\norm{ \sum_{j=0}^\infty \sum_k (d_{jk}-d'_{jk}) \psi_{jk} (X-X') } <
\end{equation}
\begin{equation}
\sum_{j=0}^\infty \sum_k \norm{d_{jk}-d'_{jk}} \psi_{jk} (X-X') \leq 
\end{equation}
\begin{equation}
\sup_{j,k} \norm{d_{jk}-d'_{jk}} = \norm{d_{j_0k}-d'_{j_0k}} = \tilde{\epsilon}
\end{equation}
Finally, the DWT is providing the dual mapping of the signal representation from the estimated detailed coefficients $d'$ within $W_j^2$ respect to the loss function $\ell_c$.
\end{proof}
The intuition behind the previous formulation is that the dual space
can create a new adversarial image by just perturbing the details
coefficients corresponding to a scale level (resolution) of the original image.
The significance of this theoretical result is that i) adversarial attacks could be further targeted to an optimal resolution (scale and frequency) and
ii) makes no assumption on the attack type since results applies to black-box, white-box and class-targeted attacks.
Next, we will demonstrate a formulation for black-box and generative models.


\subsubsection{Wavelet-based Generative Adversarial Perturbations}
Poursaeed~\textit{et al.}~\cite{poursaeed_generative_2018} proposed a Generative Adversarial Perturbation (GAP) 
method to minimize cross entropy loss, targeting the least likely class for an input image and surrogate model $c$. 
Here, noise (from uniform distribution) is passed to a 
generator network to create a perturbation. 
Then, the output of the generator is normalized based on a maximum magnitude of perturbation 
and added to the original image to generate $X'$.

\textbf{Time Domain Formulation:} We formulate a novel universal loss function for universal black-box attacks: we do not use noise models and only use
a surrogate model $c$. 
We propose a conditional loss function based on a budget parameter. This parameter controls the magnitude of the image perturbation. 
Comparing to normal cross entropy loss, the key difference is that we optimize the attack rate (fooling ratio) while keeping the amount of perturbation within certain budget $\epsilon$.
If perturbations exceeds the budget $\epsilon$, an additional penalty term will be added to the total loss. 
The same concept can be applied to gradient-based methods and targeted attacks. 
For a given budget $\epsilon$, the time domain representation of the loss function is:
\begin{equation}
\small{
\ell_c=
\begin{cases}
     \mathcal{H} \big( p( X' ) , \mathbb{I}_{p_{ll(X)}} \big )    + l * \mathcal{D}(X',X),  & \text{if } \norm{ X - X'} > \epsilon \\
     \mathcal{H} \big( p( X' ) , \mathbb{I}_{p_{ll(X)}} \big )                 & \text{otherwise}
\end{cases}
}
\label{WGAPLOSS}
\end{equation}
The first term $\mathcal{H}$ is the cross-entropy operator,
where $ \mathbb{I}_y$ denotes the one-hot encoding for
the class label $y$, probability $p( X' )$ is output probability for perturbed sample,
and $p_{ll(X)} = \argmin{ p(X) } $ is the least likely class given the original image $X$ based on model $c$.
For the second term, $l$ is a parameter to regulate the loss, $\mathcal{D}$ the structural dissimilarly:
\begin{equation}\label{SSIM}
\mathcal{D}(X',X) = 1 - {\frac {(2\mu _{X'}\mu _{X}+c_{1})(2\sigma _{X'X}+c_{2})}{(\mu _{X'}^{2}+\mu _{X}^{2}+c_{1})(\sigma _{X'}^{2}+\sigma _{X}^{2}+c_{2})}},
\end{equation}
where $\mu_X'$, $\sigma_X'$ represent the mean and covariance for $X'$ (similarly for $X$), $\sigma _{X'X}$ 
is the covariance of $X'$ and $X$. The values for $c_{1}$, $c_{2}$ stabilize the division with weak denominator.

\begin{figure*}[t]
  \begin{center}
    \includegraphics[width=0.95\linewidth]{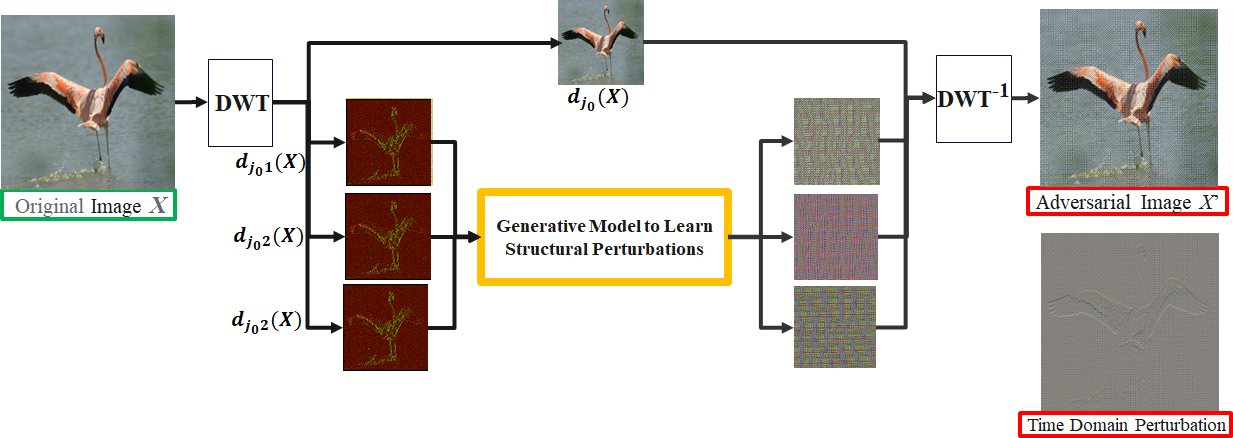}
  \end{center}
   \caption{Overview of the proposed adversarial attack.
   }
    \label{MethodAdversarialGenerative}
\end{figure*}

\textbf{Dual Formulation:} We reformulate the previous time-domain attack. From Theorem 1, given a 
budget $\epsilon$, there is a scale $j_0$ so that $X=\text{DWT}^{-1}(a_{j_0}, d_{j_0k})$ and $X'=\text{DWT}^{-1}(a_{j_0}, d'_{j_0k})$.
The, the terms for Equation~\ref{WGAPLOSS} from the following terms: 
\begin{equation}
 \mathcal{H}\big(p( \text{DWT}^{-1}(a_{j_0}, d'_{j_0k}), \mathbb{I}_{p_{ll(\text{DWT}^{-1}(a_{j_0}, d_{j_0k}))}} \big ),
\end{equation}
and
\begin{equation}
 \mathcal{D}( \text{DWT}^{-1}(a_{j_0}, d'_{j_0k}), \text{DWT}^{-1}(a_{j_0}, d_{j_0k})   ) ,
\end{equation}
where, $k \in\{1,2,3\}$. Similarly for the budget $\epsilon$ is estimated from $\text{DWT}^{-1}$. 
We show how we can use the notion of scale to constraint the magnitude of the perturbation (budget). In our implementation, we fixed a scale $j_0$, and we apply DWT to obtain the approximation
and detail coefficients  $a_{j_0}$ and $d_{j_0k}$. We only pass the detail
coefficients $d_{j_0k}$ to the generator and we reconstruct $X'$ using inverse DWT from the original approximation coefficients $a_{j_0}$ and the new perturbed coefficients $d'_{j_0k}$. Note that we are inherently normalizing the perturbed coefficients relative to the scale $j_0$ and this is possible because that approximation coefficients contain most of the energy. 
Figure~\ref{MethodAdversarialGenerative} shows a schematic of the proposed method.
\section{Results}
\label{SecResults}
We demonstrate model transferability of our Wavelet-based Generative Adversarial Perturbation (WGAP) method  and compare
performance with the generative model GAP~\cite{poursaeed_generative_2018} and a gradient-based method~\cite{cheng_improving_2019}.
The ILSVRC12 (ImageNet-1000) dataset~\cite{russakovsky_imagenet_2015} was used for training, 
and 1000 randomly selected ImageNet images~\cite{cheng_improving_2019} were used for evaluation.
The Generative model for GAP~\cite{poursaeed_generative_2018} and WGAP is a ResNet generator with 6 ResNet blocks,
except the size of the input and output channels are different. The surrogate model is Inception-v3.
To implement the discrete wavelet transform (DWT),
we used the Daubechies wavelet filter bank db2~\cite{daubechies_wavelet_1990} at scale levels: 1, 2, 3 and
budgets $\epsilon_1=0.05$, $\epsilon_2=0.1$, $\epsilon_3=0.25$.
The selection of the Daubechies wavelet is due to stable numerical approximation and compact representation.
For the WGAP models, nine input and output channels were used due to three concatenated detail coefficients.
For the first convolution layer of the generative model,
generative filter kernel size was set to 7 with a kernel number of 64.
For the second and third convolution layers,
the kernel size was 3 with a kernel number of 128.
For each of the ResNet block, the kernel size as set to 3 a kernel number of 256.
Batch normalization and ReLu activation function was used along with each convolution layer.
The Adam optimization function was used with learning rate 0.0002 and beta parameters 0.5 and 0.999.
The GAP and WGAP models were trained for 100 epochs with 500 iterations each. For GAP, we used magnitude value of 10.
The models were then evaluated using a conventional NVIDIA V100-32G GPU and implemented in Pytorch. Pretrained are from the TorchVision library and~\text{https://github.com/thu-ml/Prior-Guided-RGF}.

\subsection{Attack and defense comparison with generative-based models}
To evaluate black-box attack transferability (no access to training data, nor the target model),
we implemented adversarial attacks using Inception-v3 as a surrogate model. Testing models
include ResNet-50~\cite{he_deep_2016}, VGG-16~\cite{simonyan_very_2015}.
To evaluate attack and defense, we use three state-of-the-art defense approaches: JPEG compression~\cite{guo_countering_2018},
Guided Denoiser~\cite{liao_defense_2018}, Randomization~\cite{xie_mitigating_2018}.

Table~\ref{TableGAPComparison} shows fooling ratio in normal and defense performance for GAP~\cite{poursaeed_generative_2018},
a modified GAP formulation with budget constraint (no wavelet integration) and WGAP.
Overall, our model WGAP achieved best performance for scale level 1 and lowest for scale level 3.
Interestingly, GAP showed best performance when
using the testing networks  ResNet-50, VGG-16 than the surrogate network Inception-v3.
Best performance was obtained by WGAP, scale 1, across all budgets.
Second-best performance was obtained by \WGAP{2}{k=1,2,3}, and similar performance was obtained with GAP and \WGAP{3}{k=1,2,3}.
It is worth noting that, for a fixed budget, there is a decrease in
fooling ratio performance from \WGAP{1}{k=1,2,3} (level 1) to \WGAP{3}{k=1,2,3} (level 3) and \GAP{k=1,2,3} (no wavelet).
This suggests that defense approaches may not be tuned to overcome attacks from targeted frequency perturbations.
\begin{table*}[h]
\begin{adjustbox}{width=\columnwidth,center}
  \centering
  \begin{tabular}{lllllll}
                     & \multicolumn{3}{c}{Attacks in Normal Models}  & \multicolumn{3}{c}{Attacks in Defense Models} \\
    Method           & Inception-v3 &  ResNet-50 &  VGG-16 & JPEG Compression &  Randomization & Guided Denoiser \\
    \midrule
    GAP                    & 69.4\%       & 91.8\%     & 90.3\%  & 13.0\%           & 14.1\%         & 1.8 \%         \\
    \midrule
    \GAP{1}       & 84.4\%       & 89.5\%     & 90.8\%  & 10.7\%           & 12.3\%         & 13.4\%          \\
    \WGAP{1}{1}   & 91.1\%       & 97.5\%     & 98.5\%  & 79.2\%           & 73.5\%         & 57.8\%          \\
    \WGAP{2}{1}   & 81.1\%       & 96.5\%     & 97.3\%  & 69.1\%           & 68.8\%         & 33.9\%          \\
    \WGAP{3}{1}   & 34.2\%       & 84.4\%     & 88.7\%  & 30.7\%           & 30.3\%         & 13.6\%          \\
    \midrule
    \GAP{2}       & 92.6\%       & 94.5\%     & 97.1\%  & 15.1\%           & 17.5\%         & 18.3\%          \\
    \WGAP{1}{2}   & 97.6\%       & 98.9\%     & 99.2\%  & 96.2\%           & 85.5\%         & 86.9\%          \\
    \WGAP{2}{2}   & 85.4\%       & 97.1\%     & 99.3\%  & 77.9\%           & 76.0\%         & 61.1\%          \\
    \WGAP{3}{2}   & 55.4\%       & 89.1\%     & 95.7\%  & 51.1\%           & 47.0\%         & 22.8\%          \\
    \midrule
    \GAP{3}       & 95.1\%       & 98.6\%     & 99.6\%  & 23.3\%           & 25.1\%         & 29.8\%          \\
    \WGAP{1}{3}   & 99.4\%       & 99.8\%     & 99.7\%  & 99.7\%           & 93.6\%         & 89.7\%          \\
    \WGAP{2}{3}  & 92.2\%       & 98.0\%     & 99.3\%  & 87.8\%           & 83.6\%         & 68.9\%          \\
    \WGAP{3}{3}  & 56.3\%       & 92.4\%     & 95.7\%  & 53.0\%           & 44.5\%         & 30.7\%          \\
    \bottomrule
  \end{tabular}
  \end{adjustbox}
  \caption{Fooling ratio performance comparison for attacks in normal and defense models. Inception-v3 is the surrogate model and the others are testing models.
   GAP is the method~\cite{poursaeed_generative_2018} with magnitude value of 10,
   WGAP for scales J1-J3 and budgets $\epsilon_1$=0.05, $\epsilon_2$=0.1, $\epsilon_3$=0.25.}
  \label{TableGAPComparison}
\end{table*}
\subsection{Attack and defense comparison for iterative (query-based) models}
We further evaluate the transferability of the WGAP model by comparing performance with the Prior-guided Random Gradient-Free (P-RGF)~\cite{cheng_improving_2019}.
We use the optimal parameter $\lambda^*$ and ResNet-152 as surrogate model~\cite{cheng_improving_2019}.
The fooling ratio values for attack in normal models Inception-v3, ResNet-50 and  VGG-16 are  91.1\%, 99.6\% and 97.7\% respectively.
Similarly, the number of iterations for the same models are: 649, 352 and 370. Fooling ratio for attacks in defense models JPEG Compression,
Randomization and Guided Denoiser are 81.1\%,  82.3\% and 89.5\% respectively. Whereas for number of iterations we obtained
2120,  1816 and 1784.
As for the number of iterations, the optimized P-RGF\textsubscript{D} achieved 99.1\% in 649 iterations and the average time required
to generate an adversarial attack was 15.15 seconds.
While, P-RGF\textsubscript{D} shows very high performance in attacks for both
normal and defense models, the computationally complexity could be highly
prohibitive to use in real time scenarios.
Experimentally, we estimated an average time to produce an adversarial example,
which was 209.21 seconds per image for P-RGF\textsubscript{D}.
The most similar performance in a defense network was in the Guided Denoiser\textsuperscript{2} with a difference of 1.4\%.
Here, the P-RGF required 1816 iterations, approximately four times greater than the average number of iterations in normal models.
It should be noted that the  P-RGF requires access to the target model to iteratively estimate the gradient.
While this may be useful in attacks where a model is available, there is a risk of generalizing and over fitting.
Next, we provide an analysis to compare universal full black-box attacks and defense.
\begin{table*}[h]
  \centering
  \begin{tabular}{llllllll}
                    & P-RGF\textsubscript{D} ($\lambda^*$) & GAP     & \WGAP{1}{1} &\WGAP{2}{1} &\WGAP{1}{2} & \WGAP{2}{2}     \\
    \small{Comdefend}   &  24.5\%                              & 28.3\%  & 65.3\%      & 77.5\%     & 62.1\%     & 69.5\%      \\
    \small{ L2 Norm (\%) }&  1.4 \%                           & 3.3\%   & 4.2\%       & 4.8 \%     & 5.1\%      & 5.4 \%      \\
    \bottomrule
  \end{tabular}
  \caption{Performance comparison when the defense Comdefend method for fooling ratio and L2 Norm percentage. }
  \label{ModelTransferComparison}
\end{table*}
\subsection{Transfer attacks and defense comparison}
We compare  model transferability from all approaches for black-box attack/defense and image error approximation in terms of the L2 norm.
We test the state-of-the-art defense method Comdefend~\cite{jia_comdefend_2019} against the generated images from the surrogate models: i)
Inception-v3 for GAP and WGAP and ii) ResNet-152 for P-RGF~\cite{cheng_improving_2019}. This will provide an independent test on how general methods could be.
Please note that for the iterative method, P-RGF we do not make any new query, we just used the generated image.
Results are presented in Table~\ref{ModelTransferComparison}.
The first row shows the fooling ratio from all models (higher is better), second row shows the estimated L2 norm (lower is better).
We note that P-RGF and GAP corresponds to the poorest performance when estimating fooling ratio: 24.5\% and 28.3\% respectively. However,
the L2 norm values 1.4\% and 3.3\% for P-RGF and GAP are the lowest. For WGAP-based fooling ratio,
the values for \WGAP{1}{1} and  \WGAP{2}{1} were 65.3\% and 77.5\% respectively, while for \WGAP{1}{2} and \WGAP{2}{2} the
fooling ratio values were 62.1\% and 69.1\% respectively. We note that while there was a significance increase in performance from
budget $\epsilon_1$=0.05 to $\epsilon_2$=0.1, there was not a significant increase in L2 norm. For WGAP, we noticed that the
L2 norm error was close to 5\% even the budget was close to 10\%. This may because the network prioritizes
perturbations with no major penalty to the budget. However, there may be significant qualitative visual differences between budgets.
We note that the fooling ratio increases as wavelet coefficients go from details to approximations. This indicates
that details (high) frequency perturbation patterns may be most meaningful when
disrupting early convolutional layers from the network.

\begin{figure*}[t]
  \begin{center}
   \subfigure[Lipstick]                               {\includegraphics[width=0.19\linewidth]{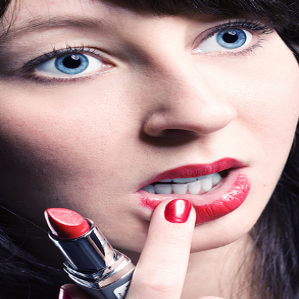}}
   \subfigure[Lipstick]                               {\includegraphics[width=0.19\linewidth]{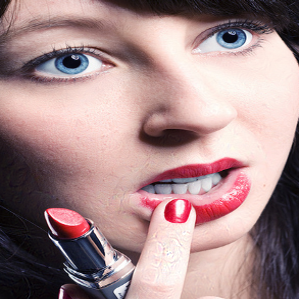}}
   \subfigure[Lipstick]                               {\includegraphics[width=0.19\linewidth]{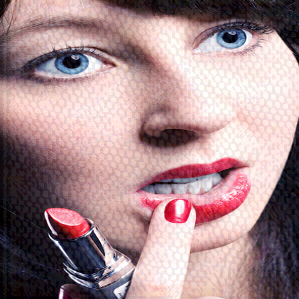}}
   \subfigure[\textcolor{red}{Thimble}]          {\includegraphics[width=0.19\linewidth]{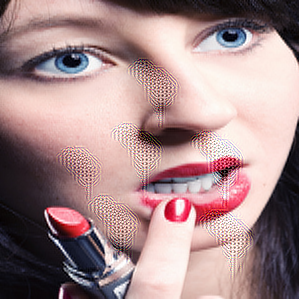}}
   \subfigure[\textcolor{red}{Bucket}]    {\includegraphics[width=0.19\linewidth]{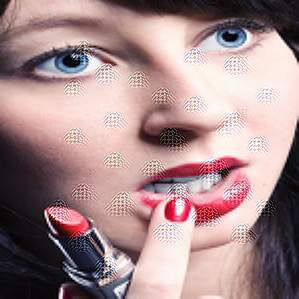}}
   \subfigure[Pizza]                            {\includegraphics[width=0.19\linewidth]{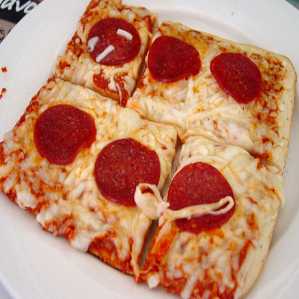} }
   \subfigure[Pizza]                            {\includegraphics[width=0.19\linewidth]{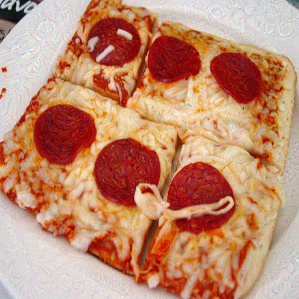}}
   \subfigure[Pizza]                            {\includegraphics[width=0.19\linewidth]{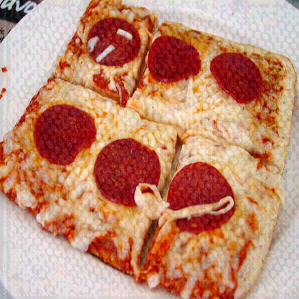} }
   \subfigure[\textcolor{red}{Screw}]     {\includegraphics[width=0.19\linewidth]{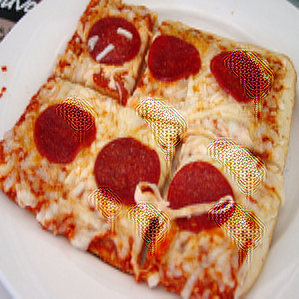} }
   \subfigure[\textcolor{red}{Rock python}]   {\includegraphics[width=0.19\linewidth]{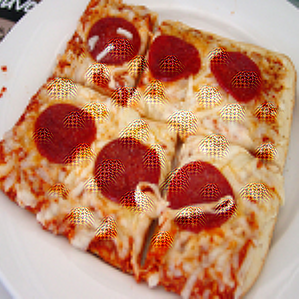}}
   \subfigure[Gasmask]                          {\includegraphics[width=0.19\linewidth]{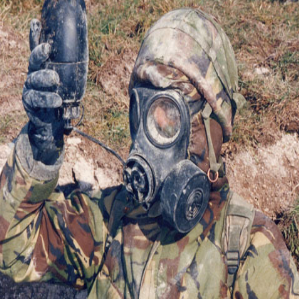}}
   \subfigure[\textcolor{red}{Oxygen mask}]      {\includegraphics[width=0.19\linewidth]{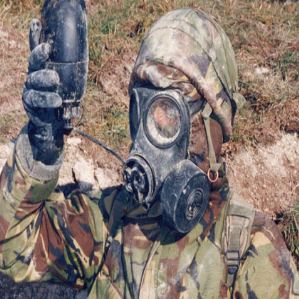}}
   \subfigure[\textcolor{red}{Comic book}]      {\includegraphics[width=0.19\linewidth]{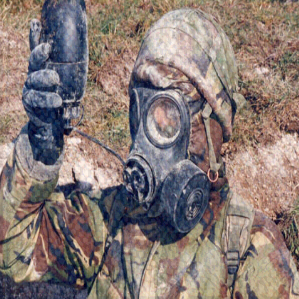}}
   \subfigure[\textcolor{red}{Strainer}]        {\includegraphics[width=0.19\linewidth]{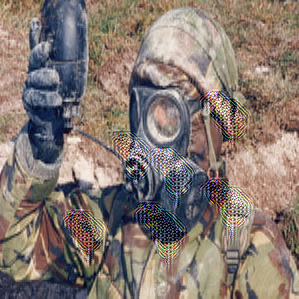}}
   \subfigure[\textcolor{red}{Zebra}]     {\includegraphics[width=0.19\linewidth]{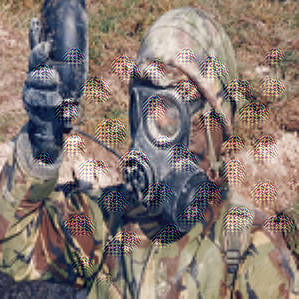}}
  \end{center}
   \caption{Adversarial examples from adversarial attack methods. 
   First column ((a),(f),(k)) corresponds to the original image .
   Second column ((b),(g),(l)) corresponds to the adversarial example generated by the P-RGF method with optimal parameter $\lambda^*$.
   Third column ((c),(h),(m)) corresponds to the GAP method, a parameter value of 10 was set as proposed in the original method.
   Fourth ((d),(i),(n)) and fifth ((e),(j),(o)) columns corresponds to our proposed method \WGAP{1}{1} and \WGAP{1}{2}, with budget $\epsilon_1=0.05$ and $\epsilon_1=0.1$ for L2 norm respectively.
   }
    \label{resultsComparison}
\end{figure*}

\begin{figure*}[t]
  \begin{center}
   \subfigure[Lipstick]                               {\includegraphics[width=0.19\linewidth]{Figs/ILSVRC2012/original_ILSVRC2012_val_00030534.png}}
   \subfigure[Lipstick]                               {\includegraphics[width=0.19\linewidth]{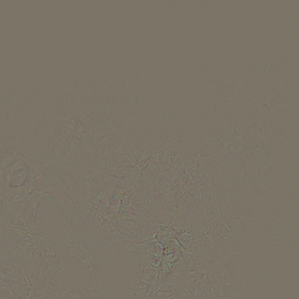}}
   \subfigure[Lipstick]                               {\includegraphics[width=0.19\linewidth]{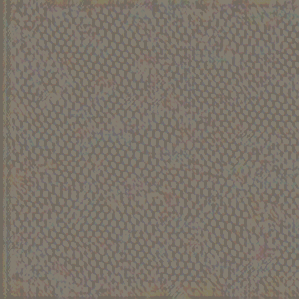}}
   \subfigure[\textcolor{red}{Thimble}]    {\includegraphics[width=0.19\linewidth]{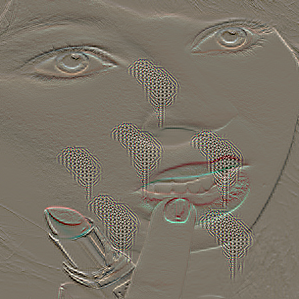}}
   \subfigure[\textcolor{red}{Bucket}]    {\includegraphics[width=0.19\linewidth]{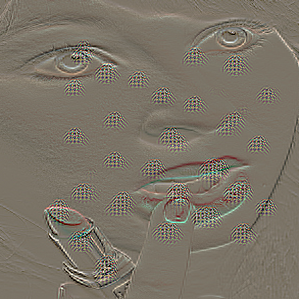}}
   \subfigure[Pizza]                            {\includegraphics[width=0.19\linewidth]{Figs/ILSVRC2012/original_ILSVRC2012_val_00015007.png} }
   \subfigure[Pizza]                            {\includegraphics[width=0.19\linewidth]{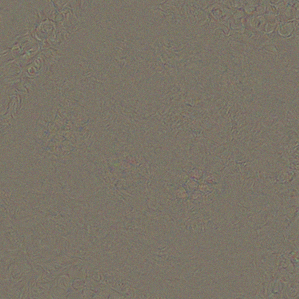}}
   \subfigure[Pizza]                            {\includegraphics[width=0.19\linewidth]{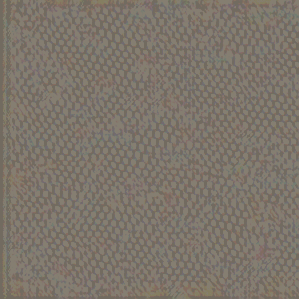} }
   \subfigure[\textcolor{red}{Screw}]     {\includegraphics[width=0.19\linewidth]{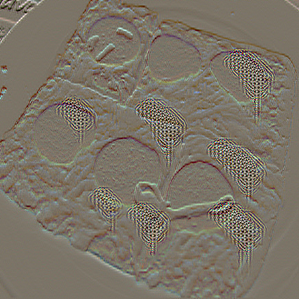} }
   \subfigure[\textcolor{red}{Rock python}]   {\includegraphics[width=0.19\linewidth]{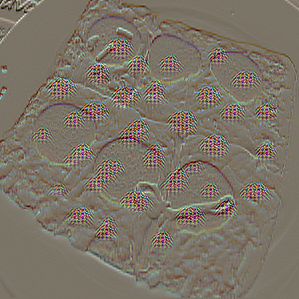}}
   \subfigure[Gasmask]                          {\includegraphics[width=0.19\linewidth]{Figs/ILSVRC2012/original_ILSVRC2012_val_00009791.png}}
   \subfigure[\textcolor{red}{Oxygen mask}]      {\includegraphics[width=0.19\linewidth]{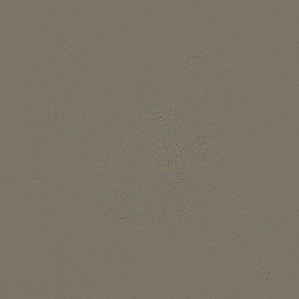}}
   \subfigure[\textcolor{red}{Comic book}]      {\includegraphics[width=0.19\linewidth]{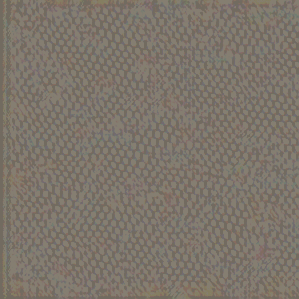}}
   \subfigure[\textcolor{red}{Strainer}]        {\includegraphics[width=0.19\linewidth]{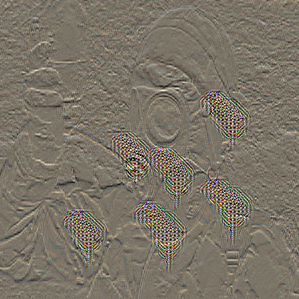}}
   \subfigure[\textcolor{red}{Zebra}]     {\includegraphics[width=0.19\linewidth]{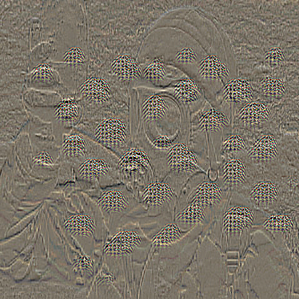}}
  \end{center}
   \caption{Residual images from adversarial attack methods. 
   First column ((a),(f),(k)) corresponds to the original image .
   Second column ((b),(g),(l)) corresponds to the adversarial example generated by the P-RGF method with optimal parameter $\lambda^*$.
   Third column ((c),(h),(m)) corresponds to the GAP method.
   Fourth ((d),(i),(n)) and fifth ((e),(j),(o)) columns corresponds to our proposed method \WGAP{1}{1} and \WGAP{1}{2}, with budget $\epsilon_1=0.05$ and $\epsilon_1=0.1$ for L2 norm respectively.
   }
    \label{resultsComparisonResidual}
\end{figure*}

Figure~\ref{resultsComparison} and Figure~\ref{resultsComparisonResidual}
depict example adversarial and residual images for each model (we omit other models due to space constraints).
Column one in Figure~\ref{resultsComparison}, shows original images with corresponding class label.
Column two and three corresponds to P-RGF~\cite{cheng_improving_2019}, and GAP~\cite{poursaeed_generative_2018} respectively.
Columns four and five corresponds to our proposed method with  \WGAP{1}{1} and \WGAP{1}{2} with scale 1 and budget $\epsilon_1$=0.05, $\epsilon_2$=0.1.

Figures~\ref{resultsComparison}(b,c) and Figures~\ref{resultsComparison}(g,h) show an example
where the defense network Comdefend, correctly classified the adversarial image. Figures~\ref{resultsComparison}(d,e) incorrectly
classified the original image, with class name Thimble and Bucket.
For Figure~\ref{resultsComparison}(l), while the defense network incorrectly classified the
original image as oxygen mask, the prediction is semantically similar to the original category (gasmask).
Figures~\ref{resultsComparison}(n,o) show the predictions from our method corresponding to Strainer and Zebra.
We notice that the most semantics difference are in relation to the budget.

Images in Figures~\ref{resultsComparisonResidual}(a,f,k) show the original images.
Images in Figures~\ref{resultsComparisonResidual}(b,g,l)  show the residual for the P-RGF method.
We can observe minimum perturbation across all images and without sign of a defined pattern across any image.
Figures~\ref{resultsComparisonResidual}(c,h,m) show the result of the perturbation when GAP method was applied.
A uniform pattern can be observed across the image. This is due to the outcome of the universal attack.
Figures~\ref{resultsComparisonResidual}(d,i,n) and Figures~\ref{resultsComparisonResidual}(e,j,o) show the
perturbations from our method at scale 1 and budgets 0.05 and 0.1, respectively. In contrast with the other two methods,
we can observe that the perturbation pattern is heterogenous and non-uniformly distributed across the image.
The perturbations is minimum at the background, either uniform (Figures~\ref{resultsComparisonResidual}(d,e) or
Figures~\ref{resultsComparisonResidual}(i,j)) or heterogeneous (Figures~\ref{resultsComparisonResidual}(n,o)).
Also, we note that the perturbations are consistently higher at the edges (as expected) and seems to increase
for larger budget. Similarly, we observe some `texture-like' patterns within the object of interest distributed across
the foreground and backgrounds which seems to adapt according to the complexity of the image texture.
This may indicate adaptive changes in the fine detail structure within the image according to the budget.
For example, in Figure~\ref{resultsComparisonResidual}(d)
there are 6 `texture-like' patterns, whereas in Figure~\ref{resultsComparisonResidual}(e) seems there are approximately 28
`texture-like' patterns distributed in the foreground. Similar `texture-like' patterns are found in the residuals
relative to the other images.
While this effect, will require more investigation, our experimental results may suggest that the
frequency spectrum is affected differently depending on the scale, budget and object of interest.
\section{Conclusions}
\label{SecDiscussion}
We presented a new formulation for  black-box attacks via time-scale (wavelets) analysis.
Our approach performs adversarial attacks where there is no access to model nor dataset  on
which the target model has been trained. We make no assumptions about the image type or the target model
and we provide extensive validation comparing our approach with two state-of-the-art adversarial attacks methods and four state-of-the-art defense algorithms.
Our results suggest that is it is both feasible and effective to  localize
coherent spatial arrangements as a mechanism to disrupt CNN layer responses.

Our effort provides the first mathematical framework in terms of image analysis and signal processing for creating attacks to understand underlying vulnerability of convolution layers in neural networks.
Even though time domain attacks are capable of effectively disrupting CNNs, it is difficult to interpret and systematically study the nature of the perturbations themselves. 
Moreover, by producing attacks from the perspective of wavelets, we seek to inspire research in academic community to look at decades of research in signal processing on wavelet transforms to study neural network properties and explore new architectures. This will in turn help us build up a general theory of how and why certain scales, orientations and frequency sub-bands affect neural networks. This will potentially lead to an understanding of CNN's vulnerability, providing the broader community with mathematical conditions to avoid unfavorable architectural choices.

To the best of our knowledge this is the first time that time-scale analysis has been
used for adversarial attacks.
The systematic analysis of the wavelet-based attacks will further provide a signal processing framework
for understanding fundamental reasons for the brittleness in state-of-the-art DL models.
The current research will be extended to provide the community with
mathematical conditions to improve DL architectures and
avoid choices that beget model fragility.

\clearpage
{\small
\bibliographystyle{ieee_fullname}
\bibliography{Universal_Attacks_CVPR_Library}
}

\end{document}